\documentclass[runningheads]{llncs}
\usepackage{cite}
\usepackage{amsmath,amssymb,amsfonts}
\usepackage{algorithmic}
\usepackage{graphicx}
\usepackage{textcomp}
\usepackage{xcolor}
\usepackage{float}
\usepackage{makecell}

\usepackage[symbol]{footmisc}

\begin{document}

\title{Iterative Self-Training for Code Generation via Reinforced Re-Ranking}

\author{
Nikita Sorokin$^{*,1}$ \and
Ivan Sedykh$^{*,1}$ \and
Valentin Malykh$^{1,2}$
}

\authorrunning{N. Sorokin, I. Sedykh, V. Malykh}

\institute{
$^{1}$ MTS AI \\
$^{2}$ International IT University \\
\email{\{n.sorokin,i.sedykh\}@mts.ai, valentin.malykh@phystech.edu}
}

\maketitle

\begin{abstract}
Generating high-quality code that solves complex programming tasks is challenging, especially with current decoder-based models that produce highly stochastic outputs. In code generation, even minor errors can easily break the entire solution. Leveraging multiple sampled solutions can significantly improve the overall output quality.

One effective way to enhance code generation is by pairing a code generation model with a reranker model, which selects the best solution from the generated samples. We propose a novel iterative self-training approach for self-training reranker models using Proximal Policy Optimization (PPO), aimed at improving both reranking accuracy and the overall code generation process. Unlike traditional PPO approaches, where the focus is on optimizing a generative model with a reward model, our approach emphasizes the development of a robust reward/reranking model. This model improves the quality of generated code through reranking and addresses problems and errors that the reward model might overlook during PPO alignment with the reranker. Our method iteratively refines the training dataset by re-evaluating outputs, identifying high-scoring negative examples, and incorporating them into the training loop, that boosting model performance.

Our evaluation on the MultiPL-E dataset demonstrates that our 13.4B parameter model outperforms a 33B model in code generation quality while being three times faster. Moreover, it achieves performance comparable to GPT-4 and surpasses it in one programming language.
\end{abstract}


\footnotetext[1]{Equal contribution.}
\footnotetext{Correspondence to: Nikita Sorokin \texttt{n.sorokin@mts.ai}}

\section{Introduction}
The generation of high-quality code using artificial intelligence models has become a critical area of research, fueled by the increasing complexity of software development tasks. Code generation models have demonstrated their potential, but traditional models, which generate code token by token, often struggle with coherence and accuracy due to the stochastic nature of their outputs. This challenge is evident in code generation benchmarks such as HumanEvalX~\cite{zheng2023codegeex} and MultiPL-E~\cite{cassano2022multiplescalableextensibleapproach}, where the disparity between Pass@1 and Pass@100 metrics can exceed two-fold.

\begin{figure}[tbh!]
\centering
\includegraphics[width=0.6\linewidth]{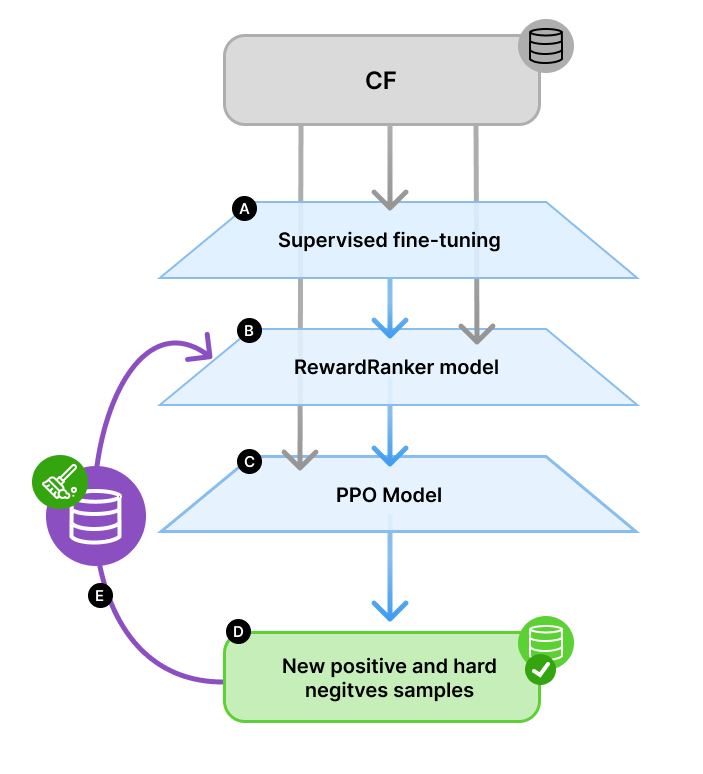}
\caption{Iterative Self-Training Workflow for RewardRanker. The process starts with supervised fine-tuning (A), followed by training the RewardRanker model (B). A PPO-based model (C) is then trained, generating new examples that are evaluated to produce both positive and hard negative samples (D). These samples are fed back into the process for further refinement and retraining (E), completing the iterative loop.}
\label{fig:yourlabel}
\end{figure}

To address these issues, we introduce RewardRanker, a reranker model designed to enhance the quality of generated code by leveraging multiple sampled solutions. Rerankers offer an effective solution to the inherent variability of code generation models by selecting the most promising candidates from a pool. This not only improves immediate output quality, but also refines the model's long-term performance through iterative self-training. We show our model training setup in Fig.~\ref{fig:yourlabel}.

Our evaluation on the MultiPL-E benchmark demonstrates that RewardRanker, with a 13.4B parameter model, outperforms larger models, including a 33B parameter model, while being three times faster. The model also achieves competitive performance with GPT-4, surpassing it in C++ programming language.
In summary, this paper introduces the following contributions:

(i) We introduce RewardRanker, a novel iterative self-training approach for reward model optimization, leveraging Proximal Policy Optimization (PPO) to refine code reranking. This iterative approach improves reranking precision and enhances code generation quality by continually fine-tuning the reward model with self-generated data.
(ii) Our approach enables smaller models (e.g., 13.4B parameters) to outperform much larger models (e.g., 33B parameters and GPT4), providing a more resource-efficient solution that maintains high code generation performance and reranking quality.
(iii) We emphasize the effectiveness of training with both correct and hard negative examples, allowing the model to generalize more robustly and improve reranking decisions across diverse coding tasks.

\section{Related Work}
The generation of source code using large language models (LLMs) has garnered significant attention, with numerous studies focusing on improving both the functional correctness and efficiency of these models. Initial experiments with models like GPT-Neo~\cite{black2022gptneox20b} and GPT-J~\cite{gpt-j} demonstrated that incorporating code into the training data enables effective program synthesis, even for medium-sized models. Concurrently, specialized models such as CodeBERT~\cite{feng-etal-2020-codebert}, GraphCodeBERT~\cite{guo_graphcodebert_2021}, Codex~\cite{chen_evaluating_2021}, CodeT5~\cite{wang_codet5_2021}, UnixCoder~\cite{guo_unixcoder_2022}, CodeGen~\cite{nijkamp2023codegen}, StarCoder~\cite{lozhkov_starcoder_2024}, phi-1~\cite{gunasekar2023textbooks}, and AlphaCode~\cite{Li_2022} have been developed to enhance code understanding and generation from natural language prompts.

Reinforcement learning (RL) approaches have also been applied to code generation. Notably, CodeRL~\cite{coderl} and PPOCoder~\cite{shojaee2023executionbased} utilize actor-critic deep RL to optimize functional correctness by testing generated code against predefined test cases. These methods highlight the potential of RL in refining code generation models to produce more accurate and executable code. Additionally, Reinforcement Learning from Human Feedback (RLHF) methodologies, including the Proximal Policy Optimization (PPO) algorithm, have been applied to align LLMs more closely with human preferences. Christiano et al. introduced RLHF, emphasizing the importance of human preferences in training models~\cite{christiano_rlhf}. This approach has been further explored and refined in various contexts, including code generation.

\cite{ni2023leverlearningverifylanguagetocode}, titled ``LEVER: Learning to Verify Language-to-Code Generation with Execution,'' presents an approach to improve language-to-code generation by training verifiers that assess the correctness of generated programs based on their execution results. Their method, combines the natural language input, the program itself, and its execution results to train verifiers that rerank sampled solutions from language models. This approach has shown significant improvements across various datasets, including table QA, math QA, and basic Python programming, with improvements ranging from 4.6\% to 10.9\% over strong baselines using execution error pruning.

Our work builds upon the foundations laid by \cite{ni2023leverlearningverifylanguagetocode}, addressing some of its limitations and extending the capabilities of verification in language-to-code generation. While LEVER focuses on reranking using execution results and verifiers, our approach, RewardRanker, leverages reinforcement learning and reward models to enhance the generation process itself. Specifically, we use a Proximal Policy Optimization (PPO) algorithm to iteratively refine our model by incorporating both correct and incorrect solutions into the learning process, which is crucial for improving code generation capabilities. Additionally, unlike LEVER, our approach does not require predefined tests during the inference stage—tests are only necessary during the training phase. This overcomes a significant limitation of the approach in \cite{ni2023leverlearningverifylanguagetocode}, making RewardRanker more flexible and scalable.

\section{Datasets}
Our research draws on a diverse dataset, primarily from CodeContests~\cite{Li_2022} and public Codeforces solutions, enriched with metadata.
For supervised fine-tuning, we structured the data into prompt-completion pairs, using problem statements as prompts and corresponding solutions as completions. We combined CodeContests and Codeforces data in equal measure, producing a balanced dataset of 1.2 million samples, each averaging 1,500 tokens. Samples longer than 4,000 tokens were excluded to optimize memory and remove outliers.

For alignment training, we formatted data as prompt-preferred-disfavored triplets. CodeContests’ mix of correct and incorrect solutions suited this structure, with “OK” verdicts marking preferred solutions (“positive”) and others as disfavored (“negative”). Triplets were formed using both minimal Levenshtein distance pairs and random incorrect-correct pairs, yielding a robust dataset of 2 million triplets.

\paragraph{Evaluation Datasets}
\textbf{MultiPL-E} \cite{cassano2022multiplescalableextensibleapproach} is a comprehensive system designed for translating unit test-driven code generation benchmarks into multiple programming languages, thus enabling multi-language code generation evaluation. MultiPL-E translates two widely-used Python benchmarks, HumanEval \cite{chen_evaluating_2021} and MBPP \cite{austin_program_2021}, into 18 additional programming languages, encompassing various paradigms and popularity levels. This system supports a diverse range of languages, making it possible to evaluate code generation models across different languages consistently.
\textbf{MBPP} (ManyBenchmarks for Python Programs) dataset, introduced by Austin et al.~\cite{austin2021programsynthesislargelanguage}, is designed for evaluating program synthesis models. It consists of 974 Python programming problems, which are sourced from real-world coding challenges and cover a variety of algorithmic tasks. Each problem includes a natural language description, a corresponding Python solution, and test cases for validation.

\section{Method}

Our approach leverages a reranker model, RewardRanker, to enhance code generation quality through an iterative self-training cycle that refines the reward model. Each cycle involves supervised fine-tuning, reward model training, and Proximal Policy Optimization (PPO) for code generation, followed by new example generation, evaluation, and retraining.

\textit{Supervised Fine-Tuning (SFT)} The generator model is initially fine-tuned on a dataset using causal language modeling to adapt to the domain. This step establishes a foundation for subsequent preference-based training, improving reranker and PPO performance.

\textit{Reward Model Training} We train a reward model to score generated code outputs, using a loss function inspired by the Bradley-Terry model to prioritize correct over incorrect solutions.

\textit{Proximal Policy Optimization (PPO)} PPO training further optimizes code generation by maximizing rewards from the reward model. Candidate solutions for tasks are generated and ranked to guide the model’s updates

\textit{Self-Training Cycle} After PPO training, the model generates new solutions, which are evaluated on task test cases. Incorrect but highly ranked solutions (hard negatives) are included in an updated training set, refining the reward model’s accuracy.

\textit{Retraining and Iterative Refinement} With an updated reward model, a new PPO model is trained from scratch, resulting in improved alignment with reward model evaluations. This cycle, while computationally costly, significantly boosts output quality even with a single iteration.

List of developed models: \textit{RewardRanker (1.3B + 6.7B)} – uses 1.3B model as code generator and 6.7B model as reranker.  \textit{RewardRanker (6.7B + 6.7B)} – both generator and reranker use 6.7B model. \textit{RewardRanker 2 iter.}$_{hardnegatives}$ – trained with hard negatives. \textit{RewardRanker 2 iter.}$_{selftraining}$ – refined with self-training examples.

\begin{table*}[tbh!]
\centering
\caption{Model performance comparison on MultiPL-E. \textbf{Best} result is in bold, \textit{second best} is in italic. Percentage of solved tasks.}
\resizebox{\textwidth}{!}{
\begin{tabular}{|c|c|c|c|c|c|c|c|c|c|c|}
\hline
\textbf{Model} & \textbf{Size} & \textbf{Python} & \textbf{C++} & \textbf{Java} & \textbf{PHP} & \textbf{TS} & \textbf{C\#} & \textbf{Bash} & \textbf{JS} & \textbf{Avg} \\
\hline
\multicolumn{10}{|c|}{Close models} \\
\hline
GPT-3.5-Turbo \cite{ye2023comprehensivecapabilityanalysisgpt3} & - & 76.2 & 63.4 & 69.2 & 60.9 & 69.1 & 70.8 & 42.4 & 67.1 & 64.9 \\
GPT-4 \cite{openai_gpt-4_2024} & - & 84.1 & 76.4 & 81.6 & 77.2 & 77.4 & 79.1 & 58.2 & 78.0 & 76.5 \\
\hline
\multicolumn{10}{|c|}{Open models} \\
\hline
DeepSeek-Coder-Instruct \cite{guo_deepseek-coder_2024} & 1.3B & 65.2 & 45.3 & 51.9 & 45.3 & 59.7 & 55.1 & 12.7 & 52.2 & 48.4 \\
DeepSeek-Coder-Instruct \cite{guo_deepseek-coder_2024} & 6.7B & 78.9 & 63.4 & 68.4 & 68.9 & 67.2 & 72.8 & 36.7 & 72.7 & 66.1 \\
DeepSeek-Coder-Instruct \cite{guo_deepseek-coder_2024} & 33B & \textit{79.3} & 68.9 & 73.4 & \textbf{72.7} & \textbf{67.9} & 74.1 & \textbf{43.0} & 73.9 & 69.2 \\
RewardRanker (1.3B + 6.7B) & 8B & 77.3 & 72.3 & 70.6 & 66.3 & 66.0 & 74.4 & 35.8 & 73.9 & 67.1 \\
RewardRanker (6.7B + 6.7B) & 13.4B & 78.9 & \textit{75.7} & \textit{74.6} & 72.1 & 66.4 & \textit{75.1} & 41.4 & \textit{74.5} & \textit{69.9} \\
RewardRanker 2 iter.$_{hardnegatives}$ & 13.4B & 80.2 & 77.9 & 73.4 & 71.6 & 66.4 & 75.8 & 38.2 & 73.8 & 69.7 \\
RewardRanker 2 iter.$_{self training}$ & 13.4B & \textbf{81.7} & \textbf{79.2} & \textbf{77.4} & \textit{71.6} & \textit{67.0} & \textbf{75.2} & \textit{39.6} & \textbf{75.1} & \textbf{70.9} \\
\hline
\end{tabular}
}
\label{tab:model_performance}
\end{table*}

\section{Results}

Our experiments validate RewardRanker’s effectiveness in improving code generation. Table \ref{tab:model_performance} shows model performance on the MultiPL-E/HumanEval-Multilingual~\cite{cassano2022multiplescalableextensibleapproach} benchmarks. For reranking, we evaluate the top 10 generated solutions per task, where DeepSeek-Coder models~\cite{guo_deepseek-coder_2024} rely on the most probable solution from top-k sampling with k=50. In contrast, RewardRanker ranks solutions based on reranker scores, yielding higher-quality outputs.

As seen in Tab.~\ref{tab:model_performance}, RewardRanker substantially improves average performance across languages. For instance, the RewardRanker 1.3B model, paired with DeepSeek-Coder-Instruct 6.7B, achieves 67.07\% accuracy, rivaling the larger 33B models. The RewardRanker 6.7B model further increases accuracy to 69.9\%, outperforming GPT-3.5-turbo while still trailing GPT-4.

The authors of \cite{ni2023leverlearningverifylanguagetocode} explored program correctness verification through execution, utilizing predefined tests to evaluate code generation outputs. However, this method relies heavily on numerous predefined test cases, which may be lacking in real-world coding tasks. To investigate this limitation, we assessed our RewardRanker model on the MBPP dataset \cite{austin2021programsynthesislargelanguage} for comparative analysis. We show the results on Tab.~\ref{tab:mbpp_results}.
While \cite{ni2023leverlearningverifylanguagetocode} reported results using \texttt{codex-davinci-002} for generation of candidates and the T5-large model \cite{2020t5} for ranking ones, we reproduced their approach with DeepSeek-Coder model~\cite{guo_deepseek-coder_2024} used in both capacities for fair comparison.
Our results demonstrate that RewardRanker consistently outperforms both variations of the LEVER approach.

\begin{table}[htb]
\centering
\caption{Model performance on MBPP}
\begin{tabular}{|c|c|c|}
\hline
\textbf{Model} & \textbf{Parameters (Billion)} & \textbf{Performance (\%)} \\
\hline
\texttt{codex-davinci-002} & - & 62.0 \\
DeepSeek-Coder-Instruct & 6.7 & 65.4 \\
Lever~\cite{ni2023leverlearningverifylanguagetocode} & - & 68.9 \\
Lever (DeepSeek-Coder) & 6.7 + 1.3 & 69.4 \\
RewardRanker & 6.7 + 1.3 & \textbf{69.9} \\
\hline
\end{tabular}
\label{tab:mbpp_results}
\end{table}

\section{Conclusion}
In this work, we introduced RewardRanker, a novel approach to enhancing code generation by combining a code generation model with a reranker model. By leveraging Proximal Policy Optimization (PPO) in an iterative self-training loop, our method refines the reranker model to improve both reranking accuracy and the overall quality of the generated code. Unlike traditional PPO approaches, which primarily focus on optimizing a generative model using a reward model, our strategy emphasizes developing a robust reward and reranking mechanism. This allows us to address and correct errors that the reward model alone might overlook.
The iterative self-training process, which incorporates hard negative examples, further enhances model performance by continuously improving the reranking of solutions. Our evaluation on the MultiPL-E dataset demonstrates the effectiveness of our approach. Our 13.4B parameter RewardRanker model outperforms larger models like a 33B parameter model.
Moreover, our model achieves performance comparable to GPT-4 and even surpasses it in one programming language.

In our future work, we plan to explore additional domains for pretraining to further investigate the combined capabilities of RewardRanker and the PPO-based generation model in a self-training setup. This will aim to improve overall quality across a broader range of coding tasks.

\bibliographystyle{splncs04}
\bibliography{lit.bib}

\end{document}